\documentclass{article}
\usepackage{arxiv}
\renewcommand{\headeright}{}
\renewcommand{\undertitle}{}
\renewcommand{\shorttitle}{}
\usepackage[utf8]{inputenc}
\usepackage[T1]{fontenc}
\usepackage{hyperref}
\usepackage{url}
\usepackage{booktabs}
\usepackage{amsfonts}
\usepackage{amsmath,amssymb,amsthm}
\usepackage{nicefrac}
\usepackage[final,babel=false]{microtype}
\usepackage{graphicx}
\usepackage{xcolor}
\usepackage{tikz}
\usetikzlibrary{arrows.meta, positioning, shapes.geometric, fit, calc, backgrounds}
\usepackage{enumitem}
\usepackage{natbib}
\setlist[itemize]{itemsep=4pt, topsep=6pt, parsep=2pt}
\setlist[enumerate]{itemsep=4pt, topsep=6pt, parsep=2pt}
\AtBeginDocument{%
  \setlength{\abovedisplayskip}{10pt plus 3pt minus 3pt}%
  \setlength{\belowdisplayskip}{10pt plus 3pt minus 3pt}%
  \setlength{\abovedisplayshortskip}{6pt plus 2pt}%
  \setlength{\belowdisplayshortskip}{6pt plus 2pt}%
}
\usepackage{etoolbox}
\makeatletter
\patchcmd{\@begintheorem}{\trivlist}{\setlength{\topsep}{8pt}\trivlist}{}{}
\makeatother
\AtBeginEnvironment{definition}{\vspace{2pt}}
\AtEndEnvironment{definition}{\vspace{2pt}}
\AtBeginEnvironment{assumption}{\vspace{2pt}}
\AtEndEnvironment{assumption}{\vspace{2pt}}
\AtBeginEnvironment{theorem}{\vspace{2pt}}
\AtEndEnvironment{theorem}{\vspace{2pt}}
\AtBeginEnvironment{lemma}{\vspace{2pt}}
\AtEndEnvironment{lemma}{\vspace{2pt}}
\AtBeginEnvironment{proposition}{\vspace{2pt}}
\AtEndEnvironment{proposition}{\vspace{2pt}}
\AtBeginEnvironment{corollary}{\vspace{2pt}}
\AtEndEnvironment{corollary}{\vspace{2pt}}
\AtBeginEnvironment{remark}{\vspace{2pt}}
\AtEndEnvironment{remark}{\vspace{2pt}}
\AtBeginEnvironment{example}{\vspace{2pt}}
\AtEndEnvironment{example}{\vspace{2pt}}
\renewcommand{\paragraph}[1]{\medskip\noindent\textbf{#1}\enspace}
\usepackage{fancyhdr}
\fancypagestyle{plain}{%
  \fancyhf{}%
  \fancyfoot[C]{\thepage}%
}
\fancypagestyle{fancy}{%
  \fancyhf{}%
  \fancyfoot[C]{\thepage}%
}
\AtBeginDocument{\thispagestyle{plain}}
\newtheorem{theorem}{Theorem}[section]

\newtheorem{proposition}[theorem]{Proposition}

\newtheorem{definition}[theorem]{Definition}
\newtheorem{example}[theorem]{Example}

\tikzset{
  block/.style={rectangle, draw, thick, minimum width=3cm, minimum height=0.8cm, align=center, fill=gray!5},
  wideblock/.style={rectangle, draw, thick, minimum width=5cm, minimum height=0.8cm, align=center, fill=gray!10},
  smallblock/.style={rectangle, draw, thick, minimum width=2.2cm, minimum height=0.7cm, align=center, fill=gray!5, font=\small},
  hub/.style={circle, draw, thick, minimum size=2cm, align=center, fill=gray!15},
  spoke/.style={rectangle, draw, thick, rounded corners, minimum width=2.4cm, minimum height=0.7cm, align=center, fill=white, font=\small},
  arr/.style={-{Stealth[length=2.5mm]}, thick},
  thinarr/.style={-{Stealth[length=2mm]}, thin}
}

\title{The AI Evaluability Gap:\\
The Missing Layer for Managing Risk and Sustaining Value}

\author{
  Vishal Srivastava \\
  Whiting School of Engineering \\
  Johns Hopkins University \\
  Baltimore, MD 21218 \\
  \texttt{vsrivas7@jh.edu} \\
  \And
  Tanmay Sah \\
  Harrisburg University of Science and Technology \\
  Harrisburg, PA \\
  \texttt{TSah@my.harrisburgu.edu} \\
}
\date{}
\begin{document}
\maketitle

% =====================================================================
\begin{abstract}
\noindent Organizations deploying AI face two fundamental challenges: they must manage AI risk (whether systems may safely operate) and sustain AI value (whether systems merit continued investment). Both rest on evidence whose sufficiency cannot be taken for granted. We name the shared underlying difficulty the \emph{Evaluability Gap}: the typical condition in which organizations lack sufficient evidence to support high-confidence governance decisions on either side. We argue this gap reflects a category error in current practice, which focuses on properties of systems rather than on the evidence required to make decisions about them, and on too narrow a conception of governance that treats investment decisions as belonging outside the governance function. Consistent with the corporate governance tradition, we argue that AI governance encompasses both decisions regarding whether a system may operate and decisions regarding whether it merits continued resources. We introduce \emph{Evaluability} as the capability of a system to generate, maintain, and renew evidence sufficient to support high-confidence governance decisions over time, and formalize governance decisions as functions of calibrated confidence $\mathrm{Conf}(D \mid E)$. We identify six properties (observability, attributability, intervenability, verifiability, calibration, and temporal validity) that an evaluable system must satisfy, and show they form a continuous cycle rather than an independent checklist. We distinguish \emph{Operational Certification} (drawing on structural evidence) from \emph{Investment Certification} (drawing on causal evidence). Evaluability does not determine whether a system is safe, fair, reliable, or valuable. It determines whether sufficient evidence exists to support high-confidence decisions regarding those properties.
\end{abstract}
\vspace{0.5em}

% =====================================================================
\section{Introduction}

Despite unprecedented investment in AI governance, organizations continue to struggle to manage AI risk and sustain AI value. The problem is often framed as a failure of controls, policies, or oversight mechanisms. We argue that a deeper issue exists: organizations frequently lack sufficient evidence to support high-confidence governance decisions. Organizations deploying AI systems at scale must manage AI risk: determining whether a system is safe, fair, reliable, and compliant enough to operate. They must also sustain AI value: determining whether the system is producing benefits that justify its continued consumption of organizational resources. Current practice often treats these as separate problems, with separate owners, separate evidence, and separate vocabularies. We argue that they are manifestations of a common underlying challenge. Both require decisions to be made under uncertainty, and both depend on evidence whose sufficiency cannot be taken for granted.

We call this shared underlying challenge the \emph{AI Evaluability Gap}: the condition in which organizations lack sufficient evidence to support high-confidence business decisions regarding either risk or value.

\begin{figure}[t]
\centering
\begin{tikzpicture}[
  topbox/.style={rectangle, draw, very thick, rounded corners, minimum width=3.2cm, minimum height=0.7cm, align=center, fill=gray!15, font=\small\bfseries},
  classbox/.style={rectangle, draw, thick, rounded corners, minimum width=5.2cm, minimum height=1.2cm, align=center, fill=gray!10, font=\small},
  domainbox/.style={rectangle, draw, thick, minimum width=5.2cm, minimum height=1.5cm, align=center, fill=white, font=\footnotesize},
  govbox/.style={rectangle, draw, thick, minimum width=6.5cm, minimum height=0.8cm, align=center, fill=gray!10, font=\small\bfseries},
  gapbox/.style={rectangle, draw=red!60, very thick, minimum width=8.5cm, minimum height=1.2cm, align=center, fill=red!8, font=\small},
  evalbox/.style={rectangle, draw, very thick, minimum width=8.5cm, minimum height=0.75cm, align=center, fill=gray!30, font=\bfseries},
  propbox/.style={rectangle, draw=none, minimum width=8.5cm, align=center, font=\footnotesize\itshape},
  chainbox/.style={rectangle, draw, thick, minimum width=6.5cm, minimum height=0.7cm, align=center, fill=white, font=\small},
  arr/.style={-{Stealth[length=2.5mm]}, thick}
]
  % Top: AI System
  \node[topbox] (sys) {AI System};
  % Two decision classes
  \node[classbox, below left=0.55cm and -1.3cm of sys] (opq) {\textbf{Operational Decisions}\\\textit{Can we operate?}};
  \node[classbox, below right=0.55cm and -1.3cm of sys] (invq) {\textbf{Investment Decisions}\\\textit{Should we continue investing?}};
  % Domain lists
  \node[domainbox, below=0.3cm of opq] (opd) {Risk \quad Fairness\\ Reliability \quad Compliance};
  \node[domainbox, below=0.3cm of invq] (invd) {Value \quad Adoption\\ Productivity \quad Business Impact};
  % Merge point
  \node[govbox, below=4.6cm of sys] (gov) {Governance Decisions};
  % The Gap
  \node[gapbox, below=0.35cm of gov] (gap) {\textbf{The Evaluability Gap}\\Insufficient evidence to support high-confidence decisions};
  % Evaluability
  \node[evalbox, below=0.35cm of gap] (eval) {Evaluability};
  % Six properties
  \node[propbox, below=0.15cm of eval] (props) {Observability \(\cdot\) Attributability \(\cdot\) Intervenability\\ Verifiability \(\cdot\) Calibration \(\cdot\) Temporal Validity};
  % Chain
  \node[chainbox, below=0.3cm of props] (chain) {Evidence \(\to\) Confidence \(\to\) Decision};
  % Arrows
  \draw[arr] (sys.south) -- ++(0,-0.15) -| (opq.north);
  \draw[arr] (sys.south) -- ++(0,-0.15) -| (invq.north);
  \draw[arr] (opq) -- (opd);
  \draw[arr] (invq) -- (invd);
  \draw[arr] (opd.south) |- ([yshift=0.4cm]gov.north) -- (gov.north);
  \draw[arr] (invd.south) |- ([yshift=0.4cm]gov.north) -- (gov.north);
  \draw[arr] (gov) -- (gap);
  \draw[arr] (gap) -- (eval);
  \draw[arr] (eval) -- (props);
  \draw[arr] (props) -- (chain);
\end{tikzpicture}
\caption{\textbf{The Evaluability Gap.} Governance of an AI system encompasses two classes of decisions: \emph{operational decisions} drawing on risk-, fairness-, reliability-, and compliance-oriented evidence, and \emph{investment decisions} drawing on value-, adoption-, productivity-, and business-impact evidence. Both feed a single governance function. The Evaluability Gap is the typical condition in which the available evidence is insufficient to warrant high-confidence governance decisions on either side. Evaluability, characterized by six properties of the evidence stream, supplies the layer that closes the gap and connects evidence to confidence to decision.}
\label{fig:gap}
\end{figure}
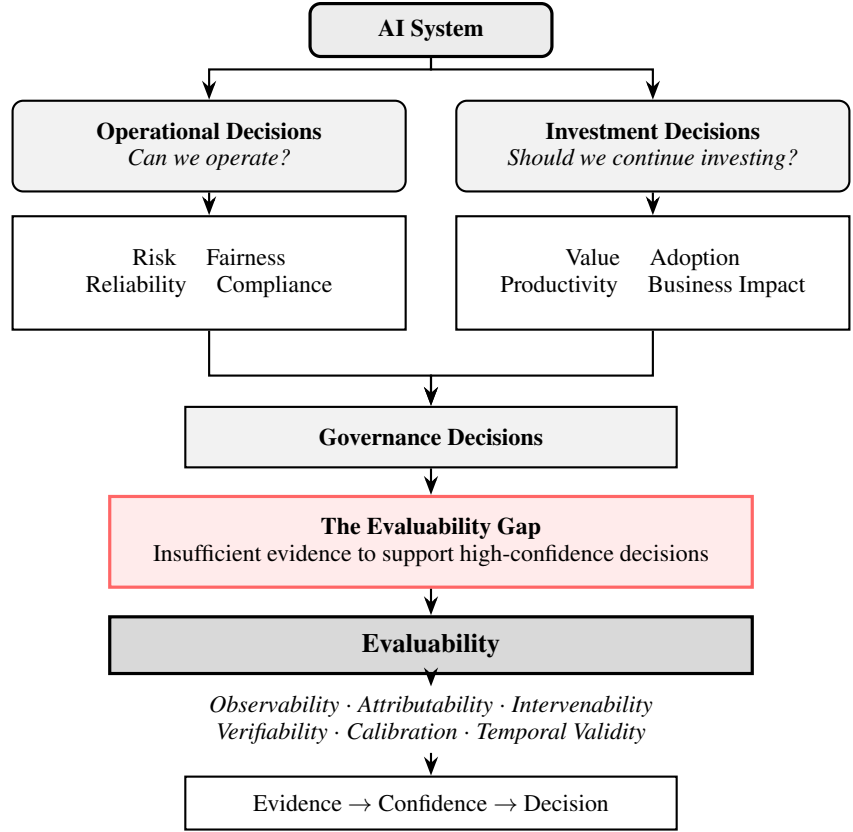

The shared underlying difficulty is what we call the \emph{Evaluability Gap} (Figure~\ref{fig:gap}).

% --- Figure 1: The Evaluability Gap (anchor figure) ---

When an organization cannot rigorously answer the question ``do we have sufficient evidence to confidently decide what to do with this system?'', it defaults either to over-deployment (accepting unmeasured risk) or to under-investment (forgoing unmeasured value). The gap is not a deficiency in any particular metric. It is the absence of an architectural commitment to the evidence required for high-confidence governance decisions, on both the operational and investment sides.

\paragraph{What governance includes.} Throughout this paper we adopt a broader conception of governance than is current in the AI assurance literature. Governance is often treated as synonymous with risk management. We adopt the broader view long established in corporate governance: boards and their delegates govern both operational matters and the allocation of organizational resources. Accordingly, governance of an AI system encompasses both the operational decision to deploy or pause and the investment decision to fund or divest. Both are governance matters because both involve organizational warrant for action under uncertainty; neither annexes the territory of strategy or finance more broadly.

We therefore define:

\begin{definition}[Evaluability]\label{def:eval}
\textbf{Evaluability} is the capability of a system to generate, maintain, and renew evidence sufficient to support high-confidence governance decisions regarding its operation and the resources allocated to it, \emph{over time}.
\end{definition}

More formally, evaluability is a function of six properties of an AI system's evidence stream:
\[
\text{Evaluability} \;=\; f(O, A, I, V, C, T),
\]
where $O$ denotes observability, $A$ attributability, $I$ intervenability, $V$ verifiability, $C$ calibration, and $T$ temporal validity (each developed in Section~\ref{sec:six-properties}). The function $f$ is intentionally unspecified. The framework requires only that $f$ be non-decreasing in each component, leaving Bayesian, Dempster--Shafer, and assurance-case-theoretic representations all admissible.

This reframing transforms the governance task. Rather than asking ``how do we measure the unmeasurable,'' it asks ``how do we architect systems so that the evidence required for any governance decision exists, is trustworthy, and remains current?''

\paragraph{Contributions.} Existing AI governance frameworks primarily specify controls, requirements, and assessment procedures. Evaluability focuses on a different question: the evidentiary conditions under which governance decisions become defensible. To our knowledge, this is the first framework that treats evidence sufficiency itself as the primary object of governance, rather than as a precondition assumed when applying domain-specific controls. The paper makes three contributions in service of this reorientation. First, we identify \emph{evidence sufficiency} as the unifying object across governance domains and formalize it through the function $\mathrm{Conf}(D \mid E)$. Second, we introduce the distinction between Operational Certification and Investment Certification (the mechanisms by which the two classes of governance decisions are warranted), clarifying the long-standing confusion about where value belongs in governance taxonomies. Third, we elevate \emph{temporal validity} from a peripheral concern to a first-class property of evidence, motivated by the distinctive non-stationarity of modern AI systems.

Evaluability is not a new theory of decision-making under uncertainty; that territory is well established \citep{savage1954, gilboa1989}. Rather, it focuses on a different question: what evidence is required before governance decisions can be made with confidence?

Its closest intellectual ancestors are the assurance-case methodologies developed for safety-critical systems \citep{kelly2004, bloomfield2010} and the audit tradition in financial reporting. Both are concerned with establishing whether available evidence is sufficient to support consequential decisions. We position Evaluability as an extension of these traditions to AI systems, where organizations must continuously determine not only whether systems may operate, but also whether they continue to justify the resources invested in them.

\section{Why Existing Governance Fails}

Before introducing the framework formally, it is useful to characterize the failure modes of current practice.

\paragraph{The silo problem.}
Organizations rarely govern AI systems through a single evidence stream. Model Risk teams build monitoring dashboards. ML Ops teams track operational performance from the deployment perspective. Business teams measure adoption and value realization. These activities often rely on overlapping data, evaluate the same system, and ultimately inform the same governance decisions, yet they operate largely independently.

The result is twofold. First, controls and instrumentation are frequently duplicated as multiple teams observe the same system for different purposes. Second, and more importantly, evidence becomes fragmented across governance functions. No single party possesses a complete view of the evidence required to support governance decisions, making it difficult to answer composite questions such as: \emph{Is this system operationally fit and delivering sufficient value to justify continued investment?}

\paragraph{The observability gap.}
Most AI systems are deployed without instrumenting the data required for counterfactual reasoning about value or for stress-testing of risk. By the time a question is asked (`did this system increase revenue?'' or `would this failure have occurred without the model in the loop?''), the evidence required to answer it has been permanently lost. Governance becomes retrospective guesswork. Where statistical answers are produced, they are typically anchored to associational rather than causal claims, leaving them vulnerable to confounding by market conditions, seasonality, and concurrent process change.

\paragraph{The non-identifiability problem.}
Even with adequate observability, the properties governance most cares about are typically not identifiable from outcomes alone. Two systems with identical observed performance may differ substantially in their causal contribution. Two systems with identical risk profiles may differ substantially in their resilience under distribution shift or susceptibility to failure.

This non-identifiability is not a measurement failure to be solved by better tooling; it is a structural feature of decisions made under uncertainty about counterfactuals and futures. Recent work establishes that for some governance-relevant properties this limitation is fundamental: information-theoretic and computational barriers may make certain safety properties undecidable from black-box outcomes alone \citep{srivastava2026limits}. 
Evaluability does not eliminate non-identifiability. Rather, it provides a disciplined framework for determining what evidence is available, what conclusions remain justified, and where uncertainty must be explicitly acknowledged. We formalize the qualitative claim as Proposition~\ref{prop:non-id} in Section~\ref{sec:framework-confidence} below.

% =====================================================================
\section{The Evaluability Framework}

We separate the theoretical layer (the logical structure of evidence-based decisions) from the institutional layer (the human bodies that produce, verify, and act on evidence). This section addresses the former; Section~\ref{sec:institutional} addresses the latter.

\subsection{Confidence as the Fundamental Object}\label{sec:framework-confidence}

Let $D$ denote a governance decision, for example ``deploy this model into production,'' ``continue funding this use case,'' or ``pause operations pending re-certification.'' Let $E$ denote the body of available evidence regarding the system, its behavior, and its context.

We define
\[
\mathrm{Conf}(D \mid E) \in [0,1]
\]
as the calibrated confidence that decision $D$ is \emph{warranted} given evidence $E$. We emphasize that $\mathrm{Conf}$ is not a probability about the world. It is an epistemic state regarding the sufficiency of evidence to support a decision under a specified standard. \emph{Confidence is not a property of the system being governed; it is a property of the evidence available to the decision maker.} The relevant lineage is legal and audit rather than ontological: ``guilty beyond reasonable doubt'' is not a probability that the defendant committed the act, but a claim that the evidence meets a specified bar.

For convenience, we also refer to the confidence associated with a governance decision as the \emph{Evaluability-Based Evidence Score} (EBES):
\[
\mathrm{EBES}(D, E) \;\equiv\; \mathrm{Conf}(D \mid E).
\]
EBES represents the degree to which the available evidence supports a particular governance decision. High EBES indicates the decision is supported by strong, trustworthy, and current evidence; low EBES indicates substantial residual uncertainty. We emphasize that in this paper EBES is a \emph{conceptual construct} rather than a prescribed measurement methodology. Future work will investigate quantitative formulations based on evidence aggregation and uncertainty reduction. We use EBES in figures and informal discussion where the focus is on evidence supporting a decision; we use $\mathrm{Conf}(D \mid E)$ in the formal framework where the focus is on the decision rule.

The governance decision rule is:
\[
\text{Take action } D \iff \mathrm{Conf}(D \mid E) > \tau(D),
\]
where $\tau(D)$ is a threshold determined by the materiality of the decision and the asymmetry of its potential errors. For human-safety-critical systems, $\tau$ is high; for low-stakes feature rollouts, $\tau$ is correspondingly lower. The framework does not prescribe specific thresholds; it requires that they be specified ex ante.

The Evaluability Gap introduced in Section~1 admits a precise statement in this notation: the Gap exists for a decision $D$ when
\[
\mathrm{Conf}(D \mid E) \;<\; \tau(D),
\]
that is, when the confidence supported by available evidence falls below the threshold required for the decision to be warranted. Governance under the Gap requires either accumulating additional evidence that raises $\mathrm{Conf}(D \mid E)$ above $\tau(D)$, or recognizing the inequality and acting accordingly: deferring the decision, applying compensating controls, or accepting that the decision cannot yet be made with warrant.

The decision-centric formulation has a structural consequence that is, in our view, the most important single implication of the framework. Because $\mathrm{Conf}$ is conditioned on $E$ rather than on observed outcomes, two systems whose observed behavior is indistinguishable need not warrant the same governance verdict.

\begin{proposition}[Governance Non-Identifiability]\label{prop:non-id}
Two AI systems may exhibit indistinguishable observed outcomes while requiring different governance decisions, because the confidence $\mathrm{Conf}(D \mid E)$ warranted by the available evidence $E$ may differ between them. Consequently, governance decisions cannot in general be inferred from outcomes alone; they depend on the structure and provenance of the evidence stream from which outcomes acquire their warrant.
\end{proposition}

Proposition~\ref{prop:non-id} is the structural reason that Evaluability, rather than observed performance, is the appropriate object of governance design. Two systems with identical accuracy can warrant different governance verdicts when one possesses the observability, attribution, and verification infrastructure that supports a high-confidence decision and the other does not. The Evaluability Gap is not closed by collecting more outcome data; it is closed by improving the evidentiary architecture from which outcome data acquire their warrant.

\begin{example}[Two Hospitals, One Model]\label{ex:hospitals}
Two hospitals deploy the same diagnostic AI system. The model is identical; on retrospective benchmarks, both report 94\% accuracy on their respective patient populations. Hospital~A deployed the system through a randomized rollout that permits causal comparison against the prior standard of care, instruments structured override logging for clinician disagreement, runs monitoring that detects drift in calibration error, and retains the audit trail required for independent verification. Hospital~B deployed the system organically: clinicians use it where they find it helpful, no comparison cohort exists, override decisions are not systematically captured, and the only available evidence is the observed accuracy figure. The two hospitals make the same observation about the system. Hospital~A can defend a high-confidence deployment decision because $\mathrm{Conf}(D \mid E_A)$ is supported by attributable, verifiable, and calibrated evidence. Hospital~B cannot, because $E_B$ contains only an observational performance figure unsupported by the architecture needed to warrant it. Performance is identifiable from outcomes; governance is not.
\end{example}

\paragraph{Representation-agnosticism.} We do not prescribe a single representation for $\mathrm{Conf}$. It may be operationalized as a Bayesian posterior, a frequency-calibrated forecast, a subjective credence supported by expert elicitation, an assurance level in the sense of \citet{kelly2004}, or a structured belief in a Dempster--Shafer framework. The framework imposes four requirements on any such representation. It must be \emph{calibrated} (stated confidences correspond to observed frequencies, where verifiable); \emph{aggregable} (evidence from multiple sources can be combined according to specified rules); \emph{auditable} (the chain from evidence to confidence is reconstructible by an independent party); and \emph{coherent under update}, in the sense that $\mathrm{Conf}(D \mid E \cup E')$ is a function of $\mathrm{Conf}(D \mid E)$ and the likelihood of new evidence $E'$ under the chosen update rule. Coherent updating ensures that strictly additive evidence consistent with prior observations does not reduce confidence, while still permitting confidence to fall when new evidence reveals problems with what was previously believed.

\paragraph{Estimation under each representation.} Each admissible representation comes with established estimation methodology, drawn from existing practice rather than introduced here. Under a Bayesian representation, $\mathrm{Conf}(D \mid E)$ is the posterior $P(D \mid E)$ computed by Bayes' rule from a prior over decisions and a likelihood model for the evidence; estimation reduces to posterior inference, and \citet{srivastava2026automation} develops a related Bayesian framework specifically for automation-risk propagation and optimal oversight in high-automation AI systems. Under a frequency-calibrated representation, $\mathrm{Conf}$ is estimated by minimizing expected calibration error on reserved test populations, by conformal prediction \citep{vovk2005}, or by backtesting against held-out outcomes. Under expert elicitation, $\mathrm{Conf}$ is produced via structured protocols such as the classical method \citep{cooke1991} and validated against proper scoring rules. Under a Dempster--Shafer representation, $\mathrm{Conf}$ is constructed from mass functions over a frame of governance hypotheses and combined via Dempster's rule. The framework's contribution is not to add a new estimator to this list; it is to specify which properties any estimator must satisfy for the decision rule $\mathrm{Conf}(D \mid E) > \tau(D)$ to be defensible. Concrete operationalization for a specific organizational context requires selecting one representation, applying its established estimation methods, and demonstrating that the result satisfies the four properties above. We treat this selection as an implementation question rather than a theoretical one.

\paragraph{Decision-centricity.} Evaluability adopts a decision-centric formulation rather than a claim-centric one. Governance ultimately acts on decisions rather than on abstract system properties. Intermediate claims regarding safety, fairness, reliability, or value may inform those decisions, but the framework treats them as decision-support artifacts rather than as the primary object of analysis. Accordingly, $\mathrm{Conf}(D \mid E)$ scores confidence that a governance action is warranted given available evidence, not confidence that a particular claim about the world is true. This choice is methodologically deliberate: it ties the framework directly to the institutional acts (deploy, pause, invest, divest) that governance functions are accountable for, and it sidesteps the philosophically vexed question of how to assign probabilities to claims whose definitions are themselves contested.

% --- Figure: Missing Layer ---
\begin{figure}[t]
\centering
\begin{tikzpicture}[node distance=0.6cm]
  \node[block] (decision) {Decision};
  \node[block, below=of decision] (conf) {Confidence};
  \node[block, below=of conf] (evid) {Evidence};
  \node[wideblock, below=of evid] (eval) {\textbf{Evaluability}};
  \draw[arr] (eval) -- (evid);
  \draw[arr] (evid) -- (conf);
  \draw[arr] (conf) -- (decision);
\end{tikzpicture}
\caption{The missing layer. Governance decisions depend on confidence; confidence depends on evidence; evidence depends on evaluability. Most current AI governance work focuses on the upper layers and treats the existence of evidence as given. We argue that the evaluability layer is where the governance burden actually sits.}
\label{fig:missing-layer}
\end{figure}
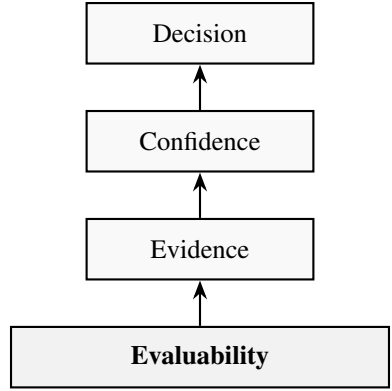

\subsection{Two Classes of Decisions: Operational and Investment Certification}\label{sec:two-classes}

Governance is often treated as synonymous with risk management. We adopt a broader view. Governance encompasses both decisions regarding whether a system may operate and decisions regarding whether it merits continued organizational investment. Operational certification addresses the former; investment certification addresses the latter. Both are pursued through the evaluability substrate developed above, but they answer different questions and rest on different evidentiary structures. We introduce the distinction here and develop its operational consequences in Section~\ref{sec:certification-detail}.

The objective of \emph{Operational Certification} is to establish operational acceptability: whether the system may operate within an envelope of acceptable risk, fairness, reliability, and compliance behavior. Its decision domains are correspondingly drawn from the operational side. The evidence required is largely \emph{structural}: specification-driven verification, stress testing, scenario analysis, adversarial evaluation. The evidentiary standard is typically high.

The objective of \emph{Investment Certification} is to establish investment justification: whether the system merits continued organizational resources, given its observed value, adoption, decision influence, productivity, and business impact. The evidence required is largely \emph{causal}: counterfactual reasoning, randomized rollout, difference-in-differences, instrumental-variable estimation. The evidentiary standard is typically lower, though it scales with the materiality of the spend.

Both certifications rest on the same evaluability substrate; what differs is the question asked, the evidence structure brought to bear, and the standard applied. The separation is what permits an organization to coherently assert that a system is operationally certified (safe and fair) but not investment-certified (no demonstrated value), or vice versa, a state which currently produces governance paralysis but which becomes actionable under the split. This is, in our view, the single most consequential implication of the framework for practice, because it gives a structured answer to a question current AI governance frameworks struggle to address: should a safe-but-valueless system continue to consume resources, and through what mechanism would that determination be made?

% --- Figure: Certification Split ---
\begin{figure}[t]
\centering
\begin{tikzpicture}[node distance=1.2cm and 1.4cm, every node/.style={align=center, font=\small}]
  \node[block] (sys) {AI System};
  \node[block, below left=1.2cm and 0.5cm of sys] (op) {Operational\\Certification};
  \node[block, below right=1.2cm and 0.5cm of sys] (inv) {Investment\\Certification};

  \node[smallblock, below=of op, yshift=0.2cm] (opd) {Risk\\Fairness\\Reliability\\Compliance};
  \node[smallblock, below=of inv, yshift=0.2cm] (invd) {Value\\Adoption\\Productivity\\Business Impact};

  \node[wideblock, below=of sys, yshift=-3.8cm] (eval) {\textbf{Evaluability}};

  \draw[arr] (sys) -- (op);
  \draw[arr] (sys) -- (inv);
  \draw[arr] (op) -- (opd);
  \draw[arr] (inv) -- (invd);
  \draw[arr] (opd) -- (eval);
  \draw[arr] (invd) -- (eval);
\end{tikzpicture}
\caption{Operational vs.\ Investment Certification. Both certifications rest on the same evaluability substrate, but they answer different questions, draw on different evidentiary structures, and apply different standards of proof. Value is not a peer to risk in the epistemic core; it is a domain of Investment Certification.}
\label{fig:cert-split}
\end{figure}
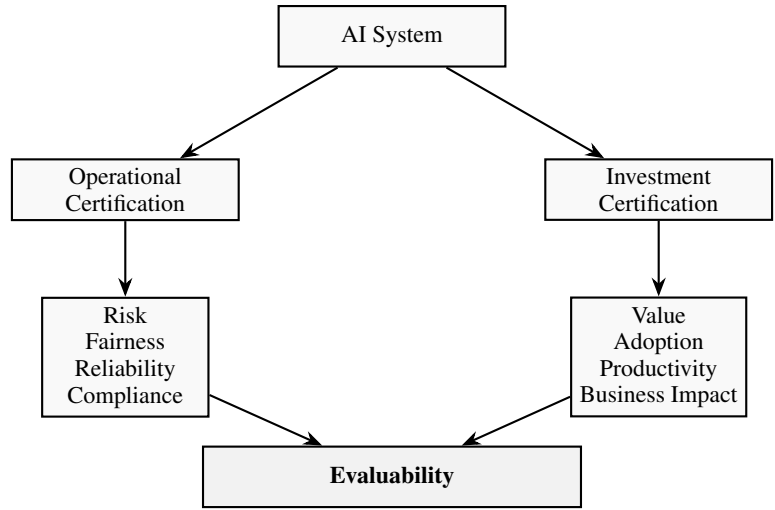

\subsection{The Six Properties of Evaluable Evidence}\label{sec:six-properties}

A system is evaluable to the extent that its evidence stream supports six properties, grouped into evidence generation and evidence validation:\footnote{An alternative grouping divides the six properties into \emph{structural} (observability, attributability, intervenability: properties of system design) and \emph{epistemic} (verifiability, calibration, temporal validity: properties of the evidence itself). We adopt the generation/validation grouping in the main text for exposition but note that the structural/epistemic alternative is equally defensible and may be preferable in design contexts.}

\paragraph{Evidence generation.}
\begin{description}
  \item[Observability.] Can the outcomes relevant to the governance decision be observed at all? Without observability, no evidence exists. Examples include logged inference inputs and outputs, downstream business outcomes (revenue, loss ratios, conversion), and user-side telemetry (override rates, time-to-decision).
  \item[Attributability.] Can the AI system's contribution to observed outcomes be isolated from confounders? Attribution requires either experimental control (randomized rollout, champion/challenger arms) or quasi-experimental design (difference-in-differences, synthetic control). Without attributability, only association can be claimed.
  \item[Intervenability.] Can the system be modified to test causal claims and to respond to failures? Intervenability includes the capacity to disable, threshold-adjust, A/B-test, and roll back. Systems that cannot be intervened on cannot be evaluated experimentally, and cannot be governed responsibly under uncertainty.
\end{description}

Generating evidence is necessary but not sufficient for governance. Organizations routinely collect vast amounts of data, logs, metrics, and reports. Yet governance failures continue to occur despite this abundance of information. The reason is that evidence must not only exist; it must also be trustworthy, interpretable, and relevant to the decision being made. Evidence validation therefore addresses a different question than evidence generation. Generation asks whether evidence can be produced. Validation asks whether that evidence should be believed.

\paragraph{Evidence validation.}
\begin{description}
  \item[Verifiability.] Can an independent party audit the evidence and reproduce the analysis? Verifiability requires logged and retained inputs, documented methodology, and access controls that permit independent inspection. Self-attested evidence is not verifiable evidence.
  \item[Calibration.] Does stated confidence correspond to observed frequencies? When the system or its overseers assert 90\% confidence, are they right 90\% of the time over the relevant reference class? Calibration is what distinguishes \emph{evidence} from \emph{trustworthy evidence}, and it is among the least-served properties in current AI governance practice.
  \item[Temporal validity.] Does the evidence remain relevant under the actual rate of environmental and capability change? An A/B test conducted on a population that no longer exists, or on a system that has since been retrained, is not current evidence. Temporal validity requires explicit decay models and renewal protocols.
\end{description}

We note that of these six, observability, attributability, intervenability, and verifiability appear in some form in existing governance frameworks. Calibration and temporal validity are the properties most underserved in current AI practice, and they bear the most weight in our framework.

Together, these six properties answer a single governance question: \emph{can this system be meaningfully evaluated?} A system may be highly accurate, highly profitable, or fully compliant, yet still fail this test if the evidence supporting those conclusions is incomplete, unverifiable, poorly calibrated, or obsolete. Evaluability therefore sits beneath traditional governance concerns. Before an organization can decide whether a system is safe, fair, reliable, or valuable, it must first determine whether those claims can be evaluated with sufficient confidence. In the notation of Definition~\ref{def:eval}, the practical implication is that $f(O, A, I, V, C, T)$ is not separable: when any single component falls below what the decision requires, the achievable evaluability is constrained regardless of how strong the remaining five are.

\paragraph{Calibration as the distinguishing property.} Of the six, calibration is the one that most sharply separates Evaluability from existing governance regimes. Most frameworks attend to whether evidence \emph{exists}: are metrics computed, monitors deployed, dashboards populated. Few attend to whether the confidence statements derived from that evidence are \emph{trustworthy}. Consider two organizations evaluating deployment risk for similar AI systems. Both report a confidence of 95\% that their system is operating safely. In the first organization, decisions labeled 95\% confidence prove correct approximately 95\% of the time. In the second, they prove correct only 65\% of the time. Although the reported confidence is identical, the evidentiary value of those statements is fundamentally different. The first organization possesses calibrated evidence; the second possesses overconfident evidence. Governance decisions based on the latter are systematically vulnerable to surprise failures.

Governance ultimately operates through confidence statements. Boards approve deployments, regulators issue certifications, and executives allocate capital because they believe the supporting evidence justifies those actions. Calibration determines whether those beliefs correspond to reality. Without calibration, governance may appear rigorous while resting on systematically distorted confidence estimates. We treat the institutional production of calibrated confidence statements (through reserved test populations, backtesting, expert elicitation under scoring rules, and conformal procedures, depending on the representation chosen) as the single most underdeveloped element of contemporary AI governance practice. A governance regime that is rich in monitoring but poor in calibration produces confident decisions whose confidence is itself unwarranted: the worst of all possible epistemic states.

\paragraph{The Evaluability Cycle.} The six properties are often interpreted as governance requirements to be checked independently. We believe this interpretation is misleading. Governance failures rarely arise because a single property is absent; they arise because weaknesses in one property propagate through the evidence lifecycle and undermine confidence elsewhere. A system with strong observability but weak attribution generates volumes of data without producing actionable evidence. A system with strong attribution but weak verification produces conclusions that cannot be independently trusted. A system with strong verification but weak temporal validity supports decisions that were correct in the past but no longer remain justified today. For this reason, the six properties should be understood as a connected system rather than a checklist.

\begin{figure}[t]
\centering
\begin{tikzpicture}[
    qbox/.style={
        rectangle, draw=blue!55!black, fill=blue!6, rounded corners=5pt,
        text width=3.6cm, align=center, minimum height=1.6cm,
        font=\sffamily\small, line width=0.6pt, inner sep=5pt,
    },
    qarrow/.style={
        -{Stealth[length=3.2mm, width=2.8mm]},
        line width=1.1pt, color=blue!55!black,
    },
    loopback/.style={
        -{Stealth[length=3.2mm, width=2.8mm]},
        line width=0.9pt, color=blue!55!black, dashed,
    },
]
% Top row (left to right)
\node[qbox] (q1) at (0, 0) {\textit{What happened?}\\[3pt]\textbf{Observability}};
\node[qbox, right=0.7cm of q1] (q2) {\textit{Did the AI cause it?}\\[3pt]\textbf{Attributability}};
\node[qbox, right=0.7cm of q2] (q3) {\textit{Can we trust the evidence?}\\[3pt]\textbf{Verifiability}};

% Bottom row (right to left)
\node[qbox, below=1.0cm of q3] (q4) {\textit{How confident should we be?}\\[3pt]\textbf{Calibration}};
\node[qbox, left=0.7cm of q4] (q5) {\textit{Does it still apply today?}\\[3pt]\textbf{Temporal Validity}};
\node[qbox, left=0.7cm of q5] (q6) {\textit{What should we do?}\\[3pt]\textbf{Intervenability}};

% Sequential arrows
\draw[qarrow] (q1) -- (q2);
\draw[qarrow] (q2) -- (q3);
\draw[qarrow] (q3) -- (q4);
\draw[qarrow] (q4) -- (q5);
\draw[qarrow] (q5) -- (q6);

% Loop-back from q6 (bottom-left) to q1 (top-left) - via the left side
\draw[loopback] (q6.west) -- ++(-0.9, 0) |- (q1.west);
\node[font=\sffamily\footnotesize\itshape, text=blue!55!black, anchor=east, align=right]
    at ($(q1.west)+(-0.95, -1.3)$) {New observations\\restart the cycle};

\end{tikzpicture}
\caption{\textbf{The Evaluability Cycle: Questions Every Governance System Must Answer.} Each property of evaluable evidence answers a distinct governance question, and together they form a continuous cycle rather than an independent checklist. Observability answers \emph{what happened?} by capturing outcomes, decisions, behaviors, and impacts. Attributability answers \emph{did the AI cause it?} through causal reasoning and experimental design. Verifiability answers \emph{can we trust the evidence?} by enabling independent inspection. Calibration answers \emph{how confident should we be?} by ensuring stated confidence corresponds to observed frequencies of correctness. Temporal validity answers \emph{does this evidence still apply today?} by assessing whether previously collected evidence remains sufficient as environments evolve. When the answer to any of these becomes uncertain, intervenability answers \emph{what should we do?} by enabling action (testing, retraining, rollback, threshold adjustment, controlled experimentation); the resulting interventions generate new observations that restart the cycle.}
\label{fig:eval-loop}
\end{figure}
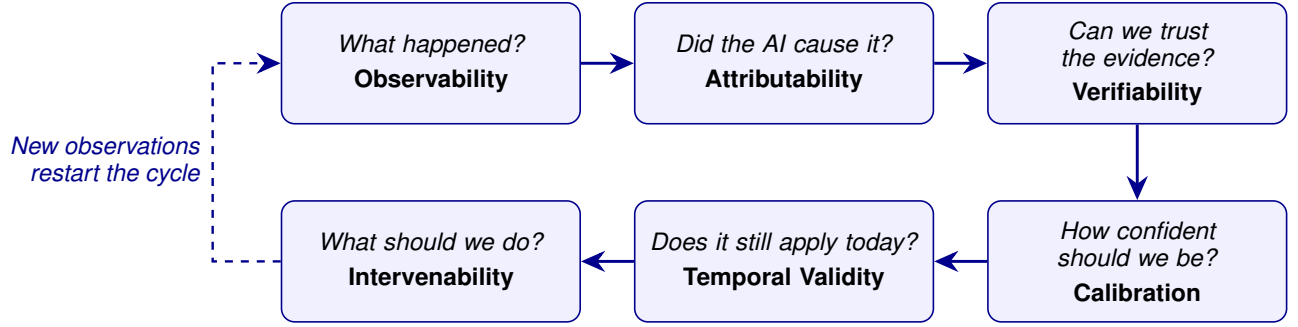

This connected system forms a continuous evidence cycle (Figure~\ref{fig:eval-loop}), and the cycle reads most intuitively as a sequence of questions that any governance regime must be able to answer. The cycle is depicted as beginning with observability because evidence cannot be generated in the absence of observable outcomes; in practice, however, it is continuous and may be entered at any stage. The first question is: \emph{what happened?} Observability captures outcomes, decisions, behaviors, and impacts as they occur. The second is: \emph{did the AI system cause it?} Attributability links observed outcomes to the system through causal reasoning and experimental design. The third is: \emph{can we trust the evidence?} Verifiability allows independent reviewers to inspect and reproduce the analysis, converting evidence from self-attested assertion into auditable artifact. The fourth is: \emph{how confident should we be?} Calibration ensures that stated confidence corresponds to observed frequencies of correctness. The fifth is: \emph{does this evidence still apply today?} Temporal validity evaluates whether previously collected evidence remains sufficient as environments, populations, regulations, and system capabilities evolve. When the answer to any of these becomes uncertain, intervenability enables governance action (testing, retraining, rollback, threshold adjustment, controlled experimentation), and these interventions generate new observations that restart the cycle. Evaluability therefore emerges not from any single property but from the sustained operation of the entire loop. Governance becomes sustainable only when organizations can repeatedly generate, validate, refresh, and act on evidence over time.

% --- Figure: The Evaluability Cycle (sequential flow with questions and properties) ---

% --- Figure: The Evaluability Cycle as a continuous capability (wheel view) ---
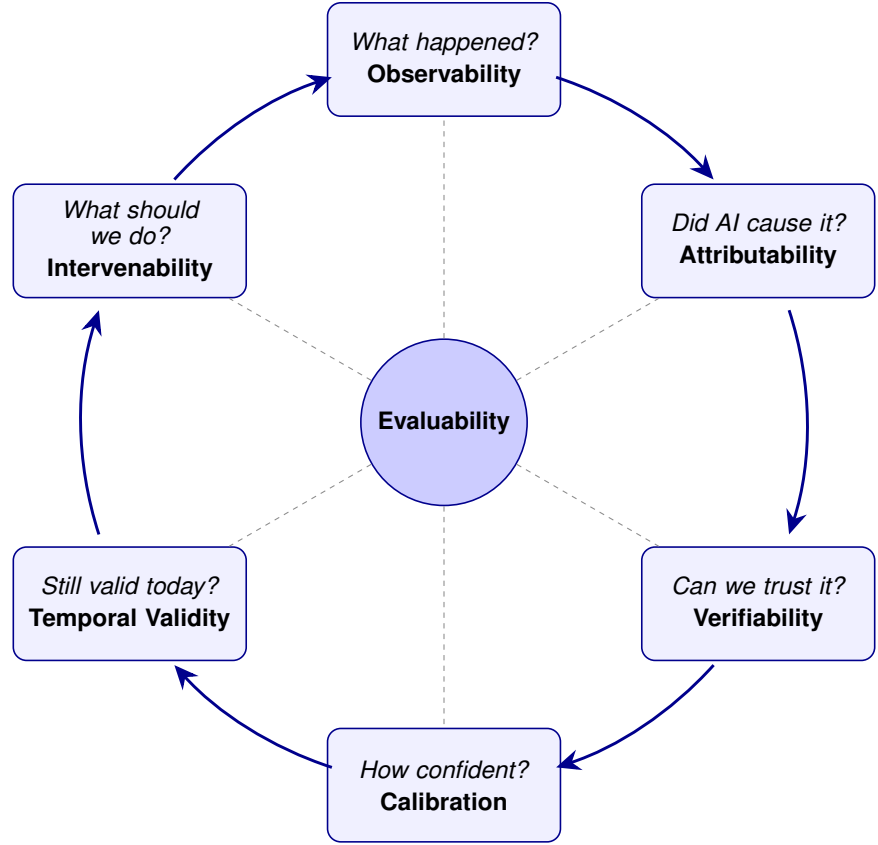
\begin{figure}[t]
\centering
\begin{tikzpicture}[
    property/.style={
        draw=blue!55!black, fill=blue!6, rounded corners=5pt,
        text width=2.8cm, align=center, minimum height=1.5cm,
        font=\sffamily\footnotesize, line width=0.6pt, inner sep=4pt,
    },
    centernode/.style={
        circle, draw=blue!55!black, fill=blue!20,
        minimum size=2.2cm, font=\sffamily\bfseries\small,
        line width=0.6pt, align=center,
    },
    cyclearrow/.style={
        -{Stealth[length=3.2mm, width=2.8mm]},
        line width=1.1pt, color=blue!55!black,
    },
    spoke/.style={
        dashed, color=black!45, line width=0.4pt,
        dash pattern=on 2pt off 2pt,
    },
]
\def\R{4.8cm}
\def\padT{18}
\def\padS{12}

\node[centernode] (eval) at (0,0) {Evaluability};

\node[property] (obs)   at (  90:\R) {\textit{What happened?}\\[2pt]\textbf{Observability}};
\node[property] (att)   at (  30:\R) {\textit{Did AI cause it?}\\[2pt]\textbf{Attributability}};
\node[property] (ver)   at ( -30:\R) {\textit{Can we trust it?}\\[2pt]\textbf{Verifiability}};
\node[property] (cal)   at ( -90:\R) {\textit{How confident?}\\[2pt]\textbf{Calibration}};
\node[property] (tem)   at (-150:\R) {\textit{Still valid today?}\\[2pt]\textbf{Temporal Validity}};
\node[property] (inter) at (-210:\R) {\textit{What should we do?}\\[2pt]\textbf{Intervenability}};

\foreach \n in {obs, att, ver, cal, tem, inter}{%
    \draw[spoke] (eval) -- (\n);
}

\foreach \angA/\padA/\angB/\padB in {%
        90/\padT/  30/\padS,
        30/\padS/ -30/\padS,
       -30/\padS/ -90/\padT,
       -90/\padT/-150/\padS,
      -150/\padS/-210/\padS,
      -210/\padS/-270/\padT}{%
    \draw[cyclearrow]
        ({\angA-\padA}:\R)
        arc[start angle={\angA-\padA}, end angle={\angB+\padB}, radius=\R];
}
\end{tikzpicture}
\caption{\textbf{Evaluability as a Continuous Capability.} The same six properties, viewed as a closed cycle rather than a sequence. The radial connections to the center indicate that Evaluability is not any single property but the joint capability of an organization to sustain all six over time. Where Figure~\ref{fig:eval-loop} emphasizes the sequence of questions a governance regime must answer, this view emphasizes that those questions are continuously re-asked: each loop around the cycle is a re-certification, and the rate at which the loop must complete depends on the system's rate of evidence decay (Section~\ref{sec:why-ai}).}
\label{fig:eval-wheel}
\end{figure}

\subsection{The Evidence Lifecycle}

Evidence is not a static asset. The lifecycle of evidence-based governance is:
\[
\text{Observation} \to \text{Evidence} \to \text{Confidence} \to \text{Decision} \to \text{Monitoring} \to \text{Updated Evidence}
\]
The loop is continuous. A system that was high-confidence at time $t_0$ may become low-confidence at time $t_1$ in the absence of renewal, not because the system has degraded, but because the evidence supporting the original decision has aged out of relevance. Evidence decays because the world changes. Users adapt, markets evolve, models are retrained, competitors respond, and regulations shift. Even when a system remains technically unchanged, the context in which it operates rarely does.

This perishability of evidence is the central motivation for treating governance not as one-time certification but as continuous re-certification. We make the dependency on time explicit. Let $E(t)$ denote the body of evidence available at time $t$. In the absence of renewal:
\[
\mathrm{Conf}\bigl(D \mid E(t)\bigr) \;\xrightarrow{t \to \infty}\; \mathrm{Conf}(D \mid \emptyset).
\]
The rate at which this decay proceeds is domain- and system-dependent. A stable linear classifier in a stationary environment decays slowly; an agentic generative system operating in a shifting market may decay rapidly. As a stylized illustration, one may write
\[
\mathrm{Conf}\bigl(D \mid E(t)\bigr) \approx \mathrm{Conf}_0 \cdot e^{-\lambda (t - t_0)},
\]
where $\lambda$ is a domain-specific decay rate.\footnote{We emphasize that this exponential form is illustrative rather than derived. Empirical evidence on model drift, concept shift, and capability evolution suggests that decay is often non-monotone and frequently discontinuous (e.g.\ regime changes, deployment of competing systems, regulatory shifts). A more faithful model would treat $\lambda$ as a stochastic process responsive to environmental signals. We retain the exponential form as a heuristic because it captures the qualitative point (evidence ages) without committing to a specific generative model.} The practical implication is concrete: re-certification cadence should be tied to the empirically estimated decay rate of the relevant evidence class, not to calendar convention.

% --- Figure: Evidence Lifecycle ---
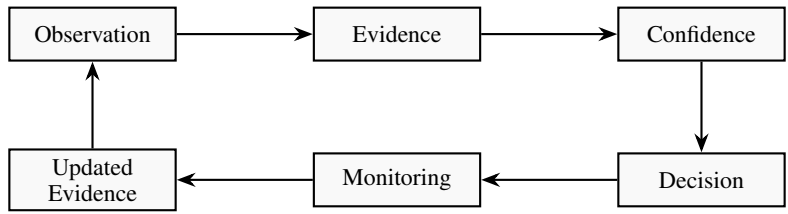
\begin{figure}[t]
\centering
\begin{tikzpicture}[node distance=1.2cm and 1.8cm]
  \node[smallblock] (obs) {Observation};
  \node[smallblock, right=of obs] (ev) {Evidence};
  \node[smallblock, right=of ev] (cf) {Confidence};
  \node[smallblock, below=of cf] (dc) {Decision};
  \node[smallblock, left=of dc] (mn) {Monitoring};
  \node[smallblock, left=of mn] (up) {Updated\\Evidence};
  \draw[arr] (obs) -- (ev);
  \draw[arr] (ev) -- (cf);
  \draw[arr] (cf) -- (dc);
  \draw[arr] (dc) -- (mn);
  \draw[arr] (mn) -- (up);
  \draw[arr] (up) |- ([yshift=-0.6cm]obs.south) -- (obs);
\end{tikzpicture}
\caption{The evidence lifecycle. Continuous re-certification is not a procedural overlay but a structural requirement: in the absence of renewal, confidence in any prior decision decays.}
\label{fig:lifecycle}
\end{figure}

% =====================================================================
\section{Why AI Makes Evaluability Hard}\label{sec:why-ai}

Evidence sufficiency is a challenge in every governed domain. Financial auditors, drug regulators, and safety engineers all confront questions of evidence quality and confidence. What makes AI distinctive is not that evidence matters, but that the evidentiary environment itself changes unusually quickly. AI systems evolve, users adapt, capabilities emerge, and operating contexts shift. Evidence that would remain valid for years in traditional domains may become obsolete in months or even weeks. Five properties of modern AI deployment make this acceleration concrete.

\paragraph{Non-stationarity.} The data on which AI systems are trained and evaluated rarely remains representative of the data on which they operate. Populations shift; competitors deploy adversarial responses; macroeconomic conditions change; user behavior adapts. Evidence collected under one distribution cannot be assumed valid under another. This is not the exception in AI deployment; it is the rule.

\paragraph{Capability emergence.} Modern foundation models exhibit capabilities that were not measured, and in some cases not anticipated, at training time. New capabilities may surface as the result of fine-tuning, prompt changes, integration with external tools, or simple usage at scale. Evaluability must contend not only with whether the system's measured behavior remains valid, but with whether the set of behaviors being measured remains the relevant set.

\paragraph{Human adaptation.} Users of AI systems modify their behavior in response to the systems they use. Underwriters who rely on a recommendation engine for ten years cease to maintain the independent judgment that the recommendation engine was originally evaluated against. Customers who interact with chatbots learn to game them. Adaptation undermines the stability of the comparison classes against which evidence is collected.

\paragraph{Opacity.} The internal reasoning of contemporary AI systems is, in most cases, not directly inspectable. Evidence about behavior must be inferred from inputs and outputs rather than read off from internal state. This places greater weight on the experimental design of attribution evidence and on the calibration of confidence assertions.

\paragraph{Agency.} Increasingly, AI systems are not passive components of a workflow but active agents that modify their environments. Agentic systems take actions, including actions that change the data-generating process against which they will subsequently be evaluated. This collapses the assumption that evaluation is something done \emph{to} a system independently of the system's behavior. \citet{srivastava2026verifier} studies a related tension for tool-using LLM agents, showing that the verification effort required to maintain safety scales unfavorably with operational horizon, exhibiting a horizon-dependent tradeoff between safety guarantees and task success: a concrete instance of the broader evaluability problem that this paper formalizes.

The combination of these five properties is what makes the temporal validity of evidence the most pressing concern in AI governance specifically. In stationary domains, evidence ages slowly. In AI, evidence ages at a rate that the framework must take seriously as a design parameter rather than an operational nuisance.

% --- Figure: Evidence Decay and Re-Certification ---
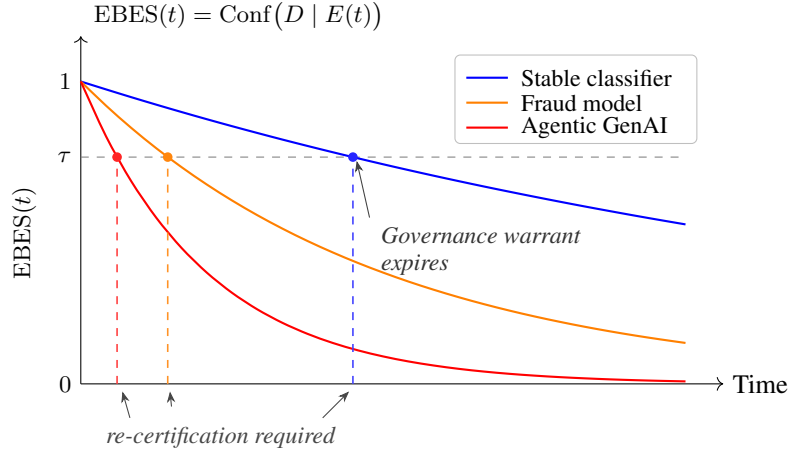
\begin{figure}[t]
\centering
\begin{tikzpicture}[scale=1]
  % Axes
  \draw[->] (0,0) -- (8.5,0) node[right] {Time};
  \draw[->] (0,0) -- (0,4.6);
  \node[anchor=east] at (0,4) {\small $1$};
  \node[anchor=east] at (0,0) {\small $0$};
  % Formal axis quantity at top
  \node[anchor=south west, font=\small] at (0.05, 4.55) {$\mathrm{EBES}(t) = \mathrm{Conf}\bigl(D \mid E(t)\bigr)$};

  % Rotated y-axis label
  \node[rotate=90, font=\small, anchor=center] at (-0.75, 2.0) {$\mathrm{EBES}(t)$};

  % Threshold line
  \draw[dashed, gray] (0,3) -- (8,3);
  \node[anchor=east] at (0,3) {\small $\tau$};

  % Three decay curves
  \draw[thick, blue, domain=0:8, smooth] plot (\x, {4*exp(-0.08*\x)});
  \draw[thick, orange, domain=0:8, smooth] plot (\x, {4*exp(-0.25*\x)});
  \draw[thick, red, domain=0:8, smooth] plot (\x, {4*exp(-0.6*\x)});

  % Re-certification markers: vertical dashed lines from x-axis up to each curve's tau crossing
  \draw[dashed, red!75, line width=0.6pt] (0.48, 0) -- (0.48, 3);
  \draw[dashed, orange!85, line width=0.6pt] (1.15, 0) -- (1.15, 3);
  \draw[dashed, blue!75, line width=0.6pt] (3.6, 0) -- (3.6, 3);

  % Dots at the crossings
  \fill[red!85] (0.48, 3) circle (1.8pt);
  \fill[orange!95] (1.15, 3) circle (1.8pt);
  \fill[blue!85] (3.6, 3) circle (1.8pt);

  % Explicit consequence label: what crossing tau actually means
  \node[font=\footnotesize\itshape, anchor=north west, text=gray!40!black, align=left] at (3.85, 2.20) {Governance warrant\\expires};
  \draw[-{Stealth[length=1.5mm]}, gray!50!black, thin] (3.82, 2.20) -- (3.65, 2.95);

  % Single below-axis annotation
  \node[anchor=north, font=\footnotesize\itshape, text=gray!55!black] at (1.85, -0.45) {re-certification required};
  \draw[-{Stealth[length=1.5mm]}, gray!55!black, thin] (0.65, -0.45) -- (0.55, -0.10);
  \draw[-{Stealth[length=1.5mm]}, gray!55!black, thin] (1.20, -0.30) -- (1.18, -0.10);
  \draw[-{Stealth[length=1.5mm]}, gray!55!black, thin] (3.20, -0.45) -- (3.55, -0.10);

  % Legend (upper-right corner; white fill so it sits cleanly over the threshold line)
  \draw[thin, gray!50, rounded corners=2pt, fill=white]
      (4.95, 3.18) rectangle (8.15, 4.35);
  \draw[thick, blue]   (5.10, 4.05) -- (5.65, 4.05);
  \node[anchor=west, font=\small] at (5.70, 4.05) {Stable classifier};
  \draw[thick, orange] (5.10, 3.72) -- (5.65, 3.72);
  \node[anchor=west, font=\small] at (5.70, 3.72) {Fraud model};
  \draw[thick, red]    (5.10, 3.39) -- (5.65, 3.39);
  \node[anchor=west, font=\small] at (5.70, 3.39) {Agentic GenAI};
\end{tikzpicture}
\caption{\textbf{Evidence Decay and the Expiration of Governance Warrant.} The vertical axis is the time-indexed Evaluability-Based Evidence Score, $\mathrm{EBES}(t) = \mathrm{Conf}(D \mid E(t))$, defined in Section~\ref{sec:framework-confidence}. The evidence supporting a governance decision decays at different rates across AI systems as users adapt, models retrain, markets evolve, and capabilities emerge; the horizontal threshold $\tau$ is the minimum EBES required for the decision to remain warranted, and the colored dots mark the moment at which each system's EBES falls below $\tau$ and re-certification becomes necessary. Under a Bayesian representation, $\mathrm{EBES}(t)$ is the time-indexed posterior; under conformal prediction it is the empirical coverage at time $t$; and similarly under the other two admissible representations surveyed in the Estimation paragraph. The decay pattern shown is structural rather than representation-specific: in the absence of renewal, $\mathrm{EBES}(t) \to \mathrm{Conf}(D \mid \emptyset)$ under any admissible representation, though the empirical decay rate $\lambda$ depends on both the system and the representation chosen. A stable classifier in a stationary environment may retain governance warrant for years; an agentic generative system in a shifting market may require re-certification on the order of weeks.}
\label{fig:decay}
\end{figure}

% =====================================================================
\section{Evidence Structure Across the Two Certifications}\label{sec:certification-detail}

We introduced the split between Operational and Investment Certification in Section~\ref{sec:two-classes} (Figure~\ref{fig:cert-split}). Here we develop its operational consequences in more depth, since the asymmetry between the two classes is not merely taxonomic: it reshapes what evidence-collection strategies are appropriate, who is positioned to produce evidence, and what counts as sufficient.

\paragraph{Asymmetry in evidence structure.} Operational evidence is, in the typical case, \emph{structural and negative-existential}. It must establish the absence of failure modes within a specified envelope: that the model does not breach a fairness threshold, does not fall below a reliability bound, does not violate compliance constraints. Such evidence is built through specification, verification, and stress testing. Investment evidence, by contrast, is \emph{causal and positive-existential}. It must establish that the system is actively producing benefit that would not have accrued in its absence. Such evidence is built through experimental and quasi-experimental designs that admit counterfactual inference. The methodologies appropriate to one are generally inappropriate to the other; A/B testing a safety claim is rarely meaningful, and specification-driven verification of value is, in most realistic deployments, simply impossible.

\paragraph{Asymmetry in evidentiary standard.} Operational standards default high because the asymmetry of error costs is acute: a wrongly-deployed unsafe system can produce harms that exceed any plausible benefit. Investment standards default lower because the cost of a wrongly-continued low-value system is the marginal spend, which is recoverable. We frame these as defaults rather than absolutes; spend-materiality tiers can and should raise the investment standard for use cases above defined thresholds, just as low-stakes operational decisions (a UI experiment touching no protected attributes) can warrant a lower bar.

\paragraph{Asymmetry in ownership.} Operational evidence is most naturally produced and verified by the parties closest to the system's technical behavior: model risk management, engineering, internal audit. Investment evidence is most naturally produced and verified by the parties closest to the business outcome: finance, product, business owners. The Evaluability framework does not collapse these into a single function; it requires that both be present and that their evidence be aggregable into a coherent overall assessment.

\paragraph{Implications for value.} Value is not an epistemic property of an AI system; it is a domain of Investment Certification. The long-running organizational confusion about who owns value measurement and what standard applies is, in our view, a direct consequence of treating value as if it were the same kind of object as risk. It is not. Conflating the two is the root cause of value claims being either over-stated by AI sponsors or dismissed by skeptics, with no shared evidentiary bar to adjudicate between them.

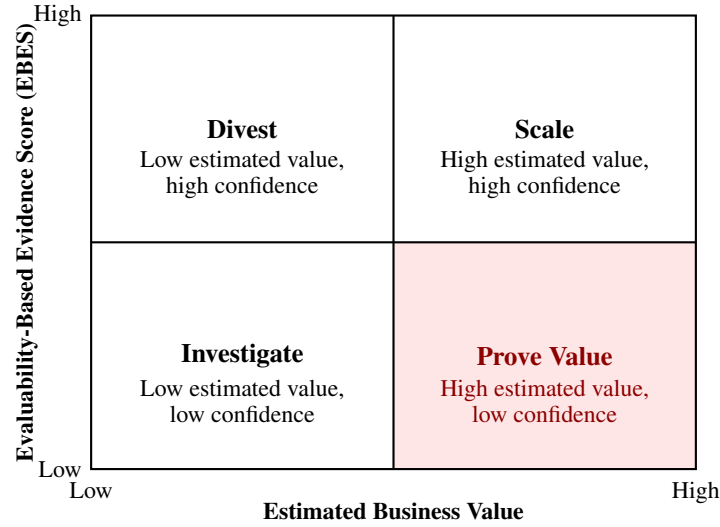
\begin{figure}[t]
\centering
\begin{tikzpicture}
  % Background shading for the dangerous quadrant (lower right)
  \fill[red!10] (4,0) rectangle (8,3);

  % Border and quadrant lines
  \draw[thick] (0,0) rectangle (8,6);
  \draw[thick] (4,0) -- (4,6);
  \draw[thick] (0,3) -- (8,3);

  % Upper-left: Divest
  \node[font=\bfseries] at (2,4.5) {Divest};
  \node[font=\small, align=center] at (2,3.9) {Low estimated value,\\high confidence};

  % Upper-right: Scale
  \node[font=\bfseries] at (6,4.5) {Scale};
  \node[font=\small, align=center] at (6,3.9) {High estimated value,\\high confidence};

  % Lower-left: Investigate
  \node[font=\bfseries] at (2,1.5) {Investigate};
  \node[font=\small, align=center] at (2,0.9) {Low estimated value,\\low confidence};

  % Lower-right: Prove Value (dangerous quadrant, emphasized)
  \node[font=\bfseries, text=red!55!black] at (6,1.5) {Prove Value};
  \node[font=\small, align=center, text=red!55!black] at (6,0.9) {High estimated value,\\low confidence};

  % Axes
  \node[rotate=90, font=\small\bfseries] at (-0.85,3) {Evaluability-Based Evidence Score (EBES)};
  \node[font=\small\bfseries] at (4,-0.55) {Estimated Business Value};
  \node[font=\small, anchor=east] at (0,6) {High};
  \node[font=\small, anchor=east] at (0,0) {Low};
  \node[font=\small, anchor=north] at (0,-0.05) {Low};
  \node[font=\small, anchor=north] at (8,-0.05) {High};
\end{tikzpicture}
\caption{\textbf{Investment Certification Requires Both Value and Evidence.} The vertical axis is the Evaluability-Based Evidence Score, $\mathrm{EBES} = \mathrm{Conf}(D \mid E)$, defined in Section~\ref{sec:framework-confidence}; for the use cases shown, $D$ is an investment decision (continue, scale, divest, investigate). Two use cases with identical estimated business value may warrant different governance actions depending on the EBES supporting those estimates: it captures whether the system has the observability, attribution, and verification infrastructure needed to warrant the value claim. The shaded lower-right quadrant is the canonical condition under which over-investment occurs (high apparent value paired with low EBES) and is the operational priority of Investment Certification: \emph{prove value} before scaling, rather than scaling on apparent value alone. The quadrant structure shown is representation-invariant: regardless of how EBES is computed (Bayesian, conformal, expert-elicited, or Dempster--Shafer; see the Estimation paragraph), two systems with identical estimated value can occupy different quadrants when the evidence supporting their value claims differs in strength, and the framework's normative claim (that the lower-right quadrant warrants investigation rather than scaling) holds across any admissible representation.}
\label{fig:portfolio}
\end{figure}

\paragraph{Two axes, not one.} The central claim of Evaluability for investment decisions is that governance must depend not only on estimated business value, but also on the confidence warranted by the supporting evidence. Two use cases with identical estimated value may require different governance actions because the strength of evidence supporting those estimates differs substantially (Figure~\ref{fig:portfolio}). The vertical axis of that figure is, in effect, a proxy for evaluability itself: it measures whether the system's observability, attribution, and verification infrastructure is strong enough to warrant the value claim. The dangerous quadrant is the one in which a use case appears valuable but lacks the evidence to justify scaling it.

% --- Figure: Investment Certification (two-axes portfolio) ---

% =====================================================================
\section{Institutional Layer: Distributed Verification}\label{sec:institutional}

A unified theory of evidence does not require, and should not produce, a centralized governance authority. To avoid a single point of epistemic capture, we distinguish three institutional roles:
\begin{itemize}[noitemsep]
  \item \textbf{Evidence Producers.} Engineering, data science, and product teams generate logs, telemetry, and experimental results in the course of building and operating systems.
  \item \textbf{Evidence Verifiers.} Model risk, internal audit, and finance functions independently validate the integrity, attribution, and calibration of that evidence.
  \item \textbf{Evidence Adjudicators.} Boards or governance committees set the decision thresholds $\tau$ and resolve conflicts between certifications, for instance when a system is operationally certified but not investment-certified.
\end{itemize}

This distribution mirrors the structure that makes financial audit work: producers, verifiers, and adjudicators are separated by design to prevent the concentration of authority. The machinery of Evaluability (the six properties, the confidence calculus, the decay model) is unified. The institutions applying that machinery should not be.

The three-role structure also has clear parallels to the three-lines-of-defense model long established in enterprise risk management, where operational management (Line 1), independent risk and compliance functions (Line 2), and internal audit (Line 3) provide layered assurance. Producers, Verifiers, and Adjudicators in our framework occupy analogous positions, with one important difference: the framework specifies what evidence flows between them, characterized by the six properties operating as a cycle, rather than only who holds which role. The institutional architecture for AI governance is not novel in its layering; it is novel in the substrate that the layers verify and adjudicate.

We deliberately limit the institutional discussion. The framework is theoretical first; operating-model questions (reporting lines, committee charters, accountability structures) depend on organizational specifics and are appropriately deferred to implementation guidance.

% =====================================================================
\section{Related Work and Positioning}

Evaluability draws on three established literatures and contributes to a fourth.

\paragraph{Assurance and safety cases.} The most direct intellectual ancestor of this framework is the assurance case methodology developed for safety-critical engineering. \citet{kelly2004} formalized the Goal Structuring Notation; \citet{bloomfield2010} provided a structured account of evidence-based safety arguments in industrial practice. Assurance cases share Evaluability's central commitment to treating safety as a claim warranted by structured evidence rather than as a property to be measured. Evaluability inherits this evidence-centered philosophy but differs in three respects. First, it extends beyond safety to encompass both operational and investment decisions, where assurance cases have historically concerned themselves with operational acceptability only. Second, it treats temporal validity and evidence decay as first-class concerns, rather than as supplementary maintenance considerations layered atop a fundamentally static argument. Third, it is designed for continuously evolving AI systems rather than predominantly static engineered systems, which has implications for the cadence of re-certification and the priority given to calibration over structural argument.

\paragraph{Audit and assurance.} The financial audit tradition provides the institutional template for distributed evidence verification, evidentiary standards graded by materiality, and continuous rather than one-time attestation. Recent work on AI assurance and verifiable claims \citep{brundage2020} has begun to import these ideas; Evaluability extends the move by making the underlying evidence substrate explicit and by separating operational from investment certification.

\paragraph{Causal inference for decision-making.} The attributability property draws directly on the counterfactual framework developed by \citet{rubin1974} and the structural causal model framework of \citet{pearl2009}. The investment certification track is, in effect, a programmatic commitment to making causal evidence first-class in AI governance.

\paragraph{AI risk-management standards.} Recent standards-based AI governance frameworks, including the NIST AI Risk Management Framework \citep{nist2023} and ISO/IEC 42001 \citep{iso2023}, specify the policies, processes, and documentation that organizations should maintain when deploying AI systems. The six properties of Evaluability map onto several of their elements: observability and attributability supply the substrate for the NIST RMF MAP and MEASURE functions; verifiability supports the GOVERN function and the audit-and-records requirements of ISO/IEC 42001; intervenability corresponds to MANAGE. Two of the six properties, calibration and temporal validity, receive comparatively little explicit treatment in current standards. The contribution of this paper to the standards-based tradition is to elevate them to first-class status and to make the evidentiary substrate beneath these largely process-oriented standards explicit, so that conformance to the standards can be assessed against the evidence required to warrant the decisions the standards are designed to govern.

\paragraph{Trustworthy AI frameworks.} Existing AI governance frameworks (including risk-management frameworks, model cards, and impact-assessment regimes) primarily specify \emph{what} should be governed: which artifacts to produce, which controls to implement, which stakeholders to consult. Evaluability focuses on a different question: the evidentiary conditions under which governance decisions become defensible. It is intended as a substrate beneath these frameworks rather than as a replacement for them. A system can be subject to model cards and still fail to be evaluable; conversely, an evaluable system supports the credible production of model cards and analogous artifacts.

% =====================================================================
\section{What Evaluability Is Not}

To prevent overreach, we delimit the framework explicitly.

Evaluability does \emph{not} determine whether a system is safe, fair, reliable, or valuable. It does not replace domain-specific risk models, fairness metrics, ROI calculators, or reliability engineering practices. It does not prescribe specific statistical methods for causal inference or specific representations of confidence.

Evaluability \emph{does} determine whether sufficient evidence exists to support decisions regarding those properties. It supplies the architectural standards (the six properties) for evidence generation and validation, the structural separation between operational and investment decisions, and the requirement that confidence be calibrated, aggregable, and auditable.

If a system has perfect Evaluability, the organization still must perform domain-specific analysis: measure the bias metric, compute the loss ratio, estimate the counterfactual value. Evaluability merely ensures that when they do, the evidence they bring to bear is trustworthy and current.

% =====================================================================
\section{Limitations}\label{sec:limitations}

Several limitations of the framework deserve explicit acknowledgment. First, Evaluability is conceptual: it has not been empirically validated against the outcomes of governance decisions in deployed enterprise settings. Second, the framework does not specify the form of $\mathrm{Conf}$, and different choices (Bayesian, frequentist, Dempster--Shafer) will yield different aggregation rules in practice. Third, and relatedly, we have deliberately left open the question of \emph{how confidence is aggregated across evidence domains}: when risk evidence, fairness evidence, reliability evidence, and value evidence all bear on the same system, the framework does not yet specify how they combine into an overall warrant for action. We regard this as the most important open theoretical question in the program and defer it to subsequent work. Fourth, the framework is silent on adversarial settings in which evidence is itself manipulable; this is a serious gap in the context of agentic AI systems whose operators may have incentives to game observability or attribution evidence. Plausible countermeasures (protected telemetry channels, independent cross-checks against operationally captured ground truth, sanctions for misreporting, and verification by parties whose incentives are not aligned with the system's continued operation) exist in mature auditing traditions, but their adaptation to AI governance is itself an open research question that we do not address here. Finally, the institutional architecture is sketched, not specified; substantial operating-model work is required to instantiate it in any specific organization.

% =====================================================================
\section{Future Work}

The framework opens a research agenda that we identify but do not pursue here.

\paragraph{Confidence aggregation.} As noted in Section~\ref{sec:limitations} above, the framework does not specify how confidence is aggregated across evidence domains. Whether the relevant aggregation rules are Bayesian (full joint posteriors), Dempster--Shafer (belief functions over a frame of discernment), or assurance-case-theoretic (structured argumentation over heterogeneous evidence) is an open question with substantial practical consequences for how composite governance decisions are reached.

\paragraph{Quantitative evaluability metrics.} The six properties are stated qualitatively. Operationalizing them requires quantitative metrics: what level of attributability is sufficient for an investment certification at a given materiality tier? What calibration error is tolerable? These thresholds will need to be derived from a combination of theoretical work and empirical study of governance outcomes.

\paragraph{Evaluability for agentic systems.} The framework as stated assumes a system whose behavior can be observed from outside. Agentic systems modify their environments and, in some cases, the conditions of their own observability. Extending Evaluability to systems whose operators may have incentives or capabilities to game the evidence stream is a non-trivial extension that the present work does not undertake.

\paragraph{Empirical validation.} The framework's claims are empirically testable: that systems built for evaluability produce better governance outcomes than those retrofitted for it; that the operational/investment split reduces governance paralysis; that calibration is the most underserved of the six properties. We see substantial value in case studies and longitudinal comparison of organizations that adopt the framework against those that do not.

% =====================================================================
\section{Conclusion}

We have argued that AI governance is, at its foundation, an evidence problem, and that the field has been working with too narrow a conception of governance. Governance does not begin and end with the determination of whether a system may operate. It also includes the determination of whether a system merits continued organizational resources. The fragmentation of current practice into risk, value, fairness, and reliability silos reflects a category error: these are not the units of governance, but rather the domains in which evidence accumulates to support two distinct classes of governance decision. By defining Evaluability as the capability of a system to generate, maintain, and renew evidence sufficient to support high-confidence decisions, we provide a unified architecture that integrates these domains under a single epistemic substrate, while preserving the distributed institutional structure that prevents capture.

The framework's central claims are these. Governance encompasses both operational and investment decisions; both require warrant, and both should be subject to the same standards of evidence sufficiency even though the evidentiary structures they draw on differ. The fundamental object of governance is confidence in a decision given evidence, not measurement of a system property. Evidence is perishable, and re-certification cadence must respond to the actual decay rate of the underlying evidence. And the properties that determine whether a system can be governed at all (observability, attributability, intervenability, verifiability, calibration, and temporal validity) must be built in at design time, not retrofitted after deployment.

Evaluability does not determine whether a system is safe, fair, reliable, or valuable. It determines whether sufficient evidence exists to support high-confidence decisions regarding those properties. As AI systems become more dynamic, adaptive, and autonomous, the ability to generate, maintain, and renew that evidence may become as important as the underlying models themselves.

% =====================================================================
\bibliographystyle{plainnat}

\end{document}